\documentclass{article} 
\usepackage{collas2022_conference,times}
\usepackage{graphicx}
\usepackage{amsmath}
\usepackage{amssymb}
\usepackage{booktabs}

\usepackage[pagebackref,breaklinks,colorlinks]{hyperref}
\usepackage{times}
\usepackage{epsfig}
\usepackage{graphicx}
\usepackage{amsmath}
\usepackage{amssymb}
\usepackage{adjustbox}
\usepackage{tabularx}
 \usepackage{verbatim}
\usepackage{bm}
\usepackage{subcaption}

 \usepackage{algorithm}

\usepackage{algorithmic}%

\usepackage{amsmath,amsfonts,bm}

\def\eqref#1{equation~\ref{#1}}

\def\1{\bm{1}}

\DeclareMathAlphabet{\mathsfit}{\encodingdefault}{\sfdefault}{m}{sl}
\SetMathAlphabet{\mathsfit}{bold}{\encodingdefault}{\sfdefault}{bx}{n}

\usepackage{hyperref}
\hypersetup{
    colorlinks=true,
    linkcolor=red,
    filecolor=magenta,      
    urlcolor=blue,
    citecolor=purple,
    pdftitle={Overleaf Example},
    pdfpagemode=FullScreen,
    }

\title{Increasing Model Generalizability for Unsupervised  Domain Adaptation}

\author{Mohammad Rostami\\
Information Sciences Institute\\
University of Southern California\\
Los Angeles, CA, USA \\
\texttt{rostamim@usc.edu} 
}

 \collasfinalcopy 

\begin{document}

\maketitle

\begin{abstract}
 A dominant approach for addressing unsupervised domain adaptation is to map data points for the source and the target domains into an embedding space which is modeled as the output-space of a shared deep encoder. The encoder is trained to make the embedding space domain-agnostic to make a source-trained classifier generalizable on the target domain. A secondary mechanism to improve UDA performance further is to make the source domain distribution more compact to improve model generalizability. We demonstrate that increasing the interclass margins in the embedding space can help to develop a UDA algorithm with improved performance. We  estimate the internally learned multi-modal distribution for the source domain, learned as a result of pretraining, and use it to increase the interclass class separation in the source domain to reduce the effect of domain shift.   We demonstrate that using our approach leads to improved model generalizability  on four standard benchmark UDA image classification datasets and compares favorably against exiting methods.
\end{abstract}

\section{Introduction}

Despite remarkable progress in deep learning, deep neural networks suffer from poor generalization when distributional gaps emerge during model execution due to the existence of \textit{domain shift}~\cite{gretton2009covariate}. Existence of distributional discrepancies between a source, i.e., training, and a target, i.e., testing, domains necessitates retraining the neural network in the target domain so the model generalizes well again. However, this process is costly and time-consuming due to requiring persistent manual data annotation~\cite{rostami2018crowdsourcing}.  
   Unsupervised Domain Adaptation (UDA) is a learning framework for mitigating the effect of domain shift when only unlabeled data is accessible in the target domain. The major approach to address UDA is to map the source domain labeled data and the target domain unlabeled data into a shared latent embedding space and then minimize the distance between the distributions in this space to mitigate domain shift~\cite{daume2009frustratingly,long2015learning,he2016deep,gabourie2019learning}. 
  When the embedding space becomes domain-agnostic, a  classifier network that receives its input from the data representations in the shared embedding space and is trained in the source domain will generalize well in the target domain.

Recent domain adaptation methods model the shared embedding space as the output-space of a deep encoder network and enforce domain alignment by training the encoder accordingly. In the UDA literature, domain alignment has been been explored extensively with most methods enforcing domain alignment either   using adversarial learning~\cite{ganin2015unsupervised,ganin2016domain,tzeng2017adversarial,hoffman2018cycada} or    through direct cross-domain distribution alignment~\cite{bhushan2018deepjdot,pan2019transferrable,rostami2021lifelong}. 
   Generative adversarial networks (GANs)~\cite{goodfellow2014generative} can be adapted to align   two distributions indirectly in an embedding space for UDA. The shared encoder is modeled as a cross-domain feature generator  network. The    source domain and the target domain features become indistinguishable  by  a competing discriminator network which is trained jointly using the adversarial min-max training procedure~\cite{goodfellow2014generative}. This procedure aligns the two distributions    effectively, but adversarial learning is known to require  delicate engineering, including  setting the optimization initial point, the architecture of the auxiliary networks, and  selection of hyper-parameters to remain stable~\cite{roth2017stabilizing} as well as mode collapse vulnerability~\cite{srivastava2017veegan}.   Direct probability matching is based on directly  minimizing a probability distribution distance  between two domain-specific distributions in the embedding space. It requires less data engineering if the right distribution metric is selected. However, performance of methods based on probability matching on average is less than methods based on adversarial learning. The reason is partially because measuring distances on higher dimensions is a challenging task, i.e., curse of dimensionality, and small distances between two   distributions in an embedding space does not necessarily translate into  semantic similarities in the input. 

Despite significant prior explorations, UDA still is an active research area.
The   general strategy of domain alignment for UDA can be improved by augmenting secondary mechanisms that increase the target domain data separability in the embedding space. A diverse set of approaches has been proposed to improve UDA performance and recent advances in UDA are primarily result of designing new secondary mechanisms. For example, Motiian et al.~\cite{motiian2017unified} enforce semantic class-consistent alignment of the distributions by assuming that a few labeled target domain data is accessible. 
Contrastive Adaptation Network~\cite{kang2019contrastive} uses pseudo-labels to enforce class-conditional alignment. Class-conditional alignment is particularly helpful to avoid matching wrong classes in the embedding space.
    Xu et al.~\cite{xu2020adversarial}  propose domain mixup on pixel and feature levels to have a continuous latent shared distribution to mitigate the oscillation of target data distribution. 
    Li et al.~\cite{li2020enhanced} regularize the loss function with entropy criterion
    to benefit from the intrinsic structure of the target domain classes. When we train a deep neural network in a supervised classification settings, data representations, i.e., network responses, form class-specific separate clusters in the final layer of the network.
A helpful strategy to improve domain alignment is to induce larger margins between these class-specific clusters in the embedding space such that the class-specific clusters become unconfined.   An approach to this end using adversarial min-max optimization to increase model generalizability~\cite{kim2019unsupervised}. In addition to be intuitive, recent theoretical results     demonstrate that large margin-separation on source  domain leads to improved model generalizability that is helpful for UDA~\cite{dhouib2020margin}.

 
{\em \bf Contributions:} we develop a new secondary mechanism to improve model generalizability to address UDA. We use the internal   data distribution that is formed in a shared embedding space, as a result of a  pretraining stage on the source domain, to increase margins between different visual class clusters  to mitigate the effect of domain shift  in a target domain. Since the internally learned distribution in the embedding space is a multimodal distribution, we use a Gaussian mixture modal (GMM) to parametrize and estimate the internal distribution. To increase the interclass margins, we build a pseudo-dataset by drawing  random samples with confident labels from the estimated GMM and regularize the model to repulse the target domain samples away from class boundaries  by minimizing the distance between the target domain and the pseudo-dataset distribution.  
We demonstrate that our algorithm is effective by  validating it on four benchmark UDA datasets  and observe that it is competitive when compared with    existing   methods.

\section{Related Work}    

We follow the direct probability matching approach for UDA. The primary challenge is to select a suitable probability discrepancy measure. A suitable metric should be computable easily and should possess nice properties for numerical minimization. Various probability metrics have been used for probability matching. To name a few works, the Maximum Mean Discrepancy (MMD) has been used for probability matching by aligning the means of the two distributions~\cite{long2015learning,long2017deep}. Sun et al.~\cite{sun2016deep} improve upon this baseline  via aligning distribution correlations to take advantage of second-order  statistics. Other improvements include using the Central moment discrepancy~\cite{zellinger2016central} and Adaptive batch normalization~\cite{li2018adaptive}. 
Domain alignment using lower order probability moments
is simple, yet it overlooks mismathces in higher moments.
The Wasserstein distance (WD)~\cite{courty2017optimal,damodaran2018deepjdot} has been used to include encoded information in higher-order statistics. 
 Damodaran et al.~\cite{damodaran2018deepjdot} demonstrated that using WD improves UDA performance compared to using MMD or correlation alignment~\cite{long2015learning,sun2016deep}.
 In contrast to more common probability metrics such as KL-divergence or JS-divergence,   WD possesses non-vanishing gradients even when the two distributions do not have overlapping supports. Hence, WD can be minimized effectively using  the first-order   optimization methods. This property makes WD a suitable choice for deep learning because deep learning objective functions are usually optimized using gradient-based methods.
  Although using WD  leads to improved UDA performance, a downside of using WD is heavy computational load in the general case compared to simpler probability metrics. The high computational load is because WD is defined as a linear programming optimization and does not have a closed-form solution for dimensions more than one.  To account for this constraint, we use the sliced Wasserstein distance (SWD)~\cite{lee2019sliced} which we have previously been used for addressing UDA in different settings~\cite{rostami2019sar,gabourie2019learning,stan2021privacy,stan2021unsupervised}. SWD is defined in terms of a closed-form solution of WD in 2D.   It can also be computed fast from empirical   samples. Compared to these works that use SWD for distributional alignment, we develop a secondary mechanism that improves model generalizability.

\begin{figure*}[t!]
    \centering
    \includegraphics[width= \linewidth]{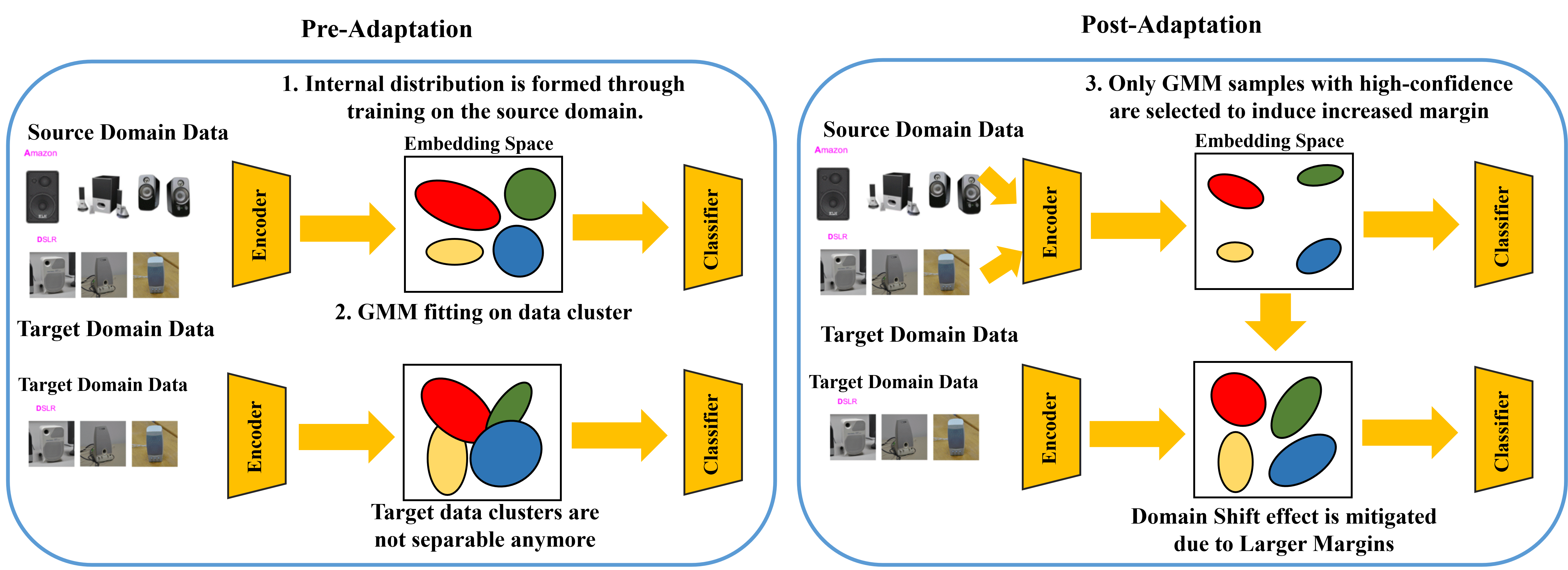}
         \caption{High-level description of the proposed unsupervised domain adaptation algorithm:    a pretrained model on the source domain learns separable source clusters in the embedding space (left-top), but the model generalizes poorly on the target domain due to domain shift which decimates class separability in the embedding space due to violating margins between data clusters (left-bottom).   The generated pseudo-dataset using confident samples helps to induce larger margins between the class clusters in the embedding space by making each class cluster more compact. As a result, we can improve generalizability (right-top) which helps to mitigate domains shift in the target domain by allowing more room for domain-shift before violating the margins between the classes (right-bottom)}
         \label{figMUDA:1}
\end{figure*}

  When a deep neural network is trained in a supervised classification   setting, learning often can be interpreted as geometric separability of the  input data points representations    in an embedding space (see Figure~\ref{figMUDA:1}). The embedding space can be
  modeled by network responses in a final higher layer.
 This interpretation implies that the input distribution is transformed into a multi-modal internal distribution, where each class is represented by a single distributional mode (see Figure~\ref{figMUDA:1}). This is necessary for class separation because final layer of a classifier network is usually a softmax layers which implies classes should become linearly separable.
 Properties for this internally learned distribution can be used to improve UDA performance. 
 For example, 
   a recent approach for UDA is to match the internal distributions   based on aligning the    cluster-specific means for each class across both domains~\cite{pan2019transferrable,chen2019progressive}. This class-aware domain alignment procedure helps to avoid class mismatch problem.   
Our goal is to regularize the shared encoder such that the internal distribution  remains robust with respect to input distributional perturbations by making class clusters more compact in the embedding space~\cite{zhang2020unsupervised}. Representation compactness will introduce large interclass margins, leading to stability with respect to domain-shift. 
Theoretical exploration in few-shot learning settings has demonstrated that increasing the inter-class margin  reduces   misclassification rate due to the existence of larger margins~\cite{cao2019theoretical}.
Inspired by these works, our idea is to estimate the formed   internal distribution as a parametric GMM and use it to induce larger margins between class-specific distributional modes.

\section{Problem Statement}  
  
Consider that we are given a target domain $\mathcal{T}$ with the unlabeled dataset $D_\mathcal{T} = (\bm{X}_\mathcal{T})$, where $\bm{X}_{\mathcal{T}}=[\bm{x}_1^t,\ldots,\bm{x}_M^t]\in\mathcal{X}\subset\mathbb{R}^{d\times M}$, denotes the input data points. The goal is to train a generalizable model for the target domain. In the absence of data annotations, this task is intractable in most cases. However, we are also given  a   source domain $\mathcal{S}$ with the labeled dataset $D_\mathcal{S} = (\bm{X}_\mathcal{S},\bm{Y}_\mathcal{S})$, where $\bm{X}_{\mathcal{S}}=[\bm{x}_1^s,\ldots,\bm{x}_N^s]\in\mathcal{X}\subset\mathbb{R}^{d\times N}$   and   $\bm{Y}_{\mathcal{S}}=[\bm{y}^s_1,...,\bm{y}^s_N]\in \mathcal{Y}\subset\mathbb{R}^{k\times N}$ denotes the corresponding one-hot labels. The two domain are related and  both share the same $k$ semantic classes. 
The source and the target input  data points are drawn i.i.d from two domain-specific distributions  $\bm{x}_i^s\sim p_{S}(\bm{x})$  and $\bm{x}_i^t\sim p_{T}(\bm{x})$. Despite cross-domain similarities, distribution discrepancy exists between the two domains, i.e, $p_{T}(\bm{x})\neq p_{S}(\bm{x})$.  In the absence of a target domain annotated dataset, a naive approach is to select  a   family of parameterized functions $f_{\theta}:\mathbb{R}^d\rightarrow \mathcal{Y}$, e.g.,  deep neural networks with learnable weight parameters $\theta$, and search for the Bayes-optimal   model   $f_{\theta^*}(\cdot)$ in this family using the standard empirical risk minimization (ERM) using the source domain annotated data: $\hat{ \theta}=\arg\min_{\theta}\{\hat{e}_{\theta}(\bm{X}_{\mathcal{S}},\bm{Y}_{\mathcal{S}},\mathcal{L})\}=\arg\min_{\theta}\{\frac{1}{N}\sum_i \mathcal{L}(f_{\theta}(\bm{x}_i^s),\bm{y}_i^s)\}$, where $\mathcal{L}(\cdot,\cdot)$ is a proper point-wise loss function, e.g., the cross-entropy loss. 
In an ideal situation, the ERM-trained model is generalizable on the source domain and due to cross-domain knowledge transfer, this source-trained model  likely would perform better than chance in the target domain. However, the source-trained model will still suffer from   generalization degradation compared to the performance on the source domain due to the existence of domain gap, i.e., $p_{T}(\bm{x})\neq p_{S}(\bm{x})$.  

The above naive solution does not benefit from the target domain unannotated dataset.
The goal in UDA is to take advantage of the encoded information  in the
unlabeled target data points and improve the source-trained  model generalization  in the target domain. We use the common approach of reducing the domain gap across the two domains by mapping data points from both domain into a shared embedding space $\mathcal{Z}$ using a shared encoder such that the distribution discrepancy becomes minimal in the embedding space. In other words, domain gap between the two domains becomes negligible after this transformation. The critical challenge is to find such an embedding space. Diversity among the many existing UDA algorithms stems from how to exactly find such a shared embedding space.

To model the above shared embedding, the end-to-end deep neural network $f_\theta(\cdot)$ can be decomposed into  an encoder subnetwork $\phi_{\bm{v}}(\cdot): \mathcal{X}\rightarrow \mathcal{Z}\subset \mathbb{R}^p$ and a classifier subnetwork $h_{\bm{w}}(\cdot): \mathcal{Z}\rightarrow \mathcal{Y}$ with learnable parameters $\bm{v}$ and $\bm{w}$, respectively. In this decomposition, we have: $f_\theta = h_{\bm{w}}\circ \phi_{\bm{v}}$ and $\theta=(\bm{w},\bm{v})$. Following the class separability condition for a good model generalization on source domain, we assume that   the classes are  separable in $\mathcal{Z}$ for the source domain as a result of supervised pretraining in the source domain (see Figure~\ref{figMUDA:1}, left).
Benefiting from the target domain data, most UDA framework adapt the source-trained encoder such that the distributions of both domains are matched in the embedding space $\mathcal{Z}$, i.e., we have $\phi(p_{\mathcal{S}}(\cdot)) \approx\phi(p_{\mathcal{T}}(\cdot))$. 
Hence, the classifier subnetwork will generalize well in the target domain, despite having been trained using solely the source domain data. 
A major class of UDA methods match the distributions $\phi(p_{\mathcal{S}}(\bm{x}^s))$ and $\phi(p_{\mathcal{T}}(\bm{x}^t))$ by training the encoder  $\phi(\cdot)$   such that the distance between these two distribution, in terms of a probability distribution metric, is minimized in the embedding space:  
 \begin{equation}
 {
\begin{split}
\hat{\bm{v}},\hat{\bm{w}}=&\arg\min_{\bm{v},\bm{w}} \frac{\lambda}{N}\sum_{i=1}^N \mathcal{L}\big(h_{\bm{w}}(\phi_{\bm{v}}(\bm{x}_i^s)),\bm{y}_i^s\big)+
 D\big(\phi_{\bm{v}}(p_{\mathcal{S}}(\bm{X}_{\mathcal{S}})),\phi_{\bm{v}}(p_{\mathcal{T}}(\bm{X}_{\mathcal{T}}))\big),
\end{split}
}
\label{eq:smallmainPrMatch}
\end{equation}  
 where $D(\cdot,\cdot)$ denotes a probability discrepancy metric and $\lambda$ is a trade-off parameter between the empirical risk and the domain alignment terms. There are various choices for $D(\cdot,\cdot)$ and many UDA methods are developed by selecting difference probablity discrepancy metrics. In this work, we select  SWD  for the metric $D(\cdot,\cdot)$.

 SWD is defined by applying the idea of slicing~\cite{le2019tree} on the  Wasserstein distance (WD), defined as:  
\begin{equation}
W(p_{\mathcal{S}},p_{\mathcal{T}})=\text{inf}_{\gamma\in \Gamma(p_{\mathcal{S}},p_{\mathcal{T}})} \int_{{X}\times {Y}} c(x,y)d\gamma(x,y)
\label{eq:kantorovich}
\end{equation}
where $\Gamma(p_{\mathcal{S}},p_{\mathcal{T}})$ denotes the set of joint distributions $p_{{\mathcal{S}},{\mathcal{T}}}$  with such that  distributions $p_{\mathcal{S}}$ and $p_{\mathcal{T}}$ are its two marginal distribution and $c:X\times Y\rightarrow \mathbb{R}^+$ is a cost function, e.g., $\ell_2$-norm Euclidean distance.
In general, WD does not have a closed form solution.  
However, in the case of $1-$dimensional  distributions,  it has a closed-form solution as follows:
\begin{equation}
W(p_{\mathcal{S}},p_{\mathcal{T}})= \int_{0}^1 c(P_{\mathcal{S}}^{-1}(\tau),P_{\mathcal{T}}^{-1}(\tau))d\tau,
\label{eq:oneD}
\end{equation} 
where $P_{\mathcal{S}}$ and $P_{\mathcal{T}}$ are the cumulative distributions of the $1-$dimensional  distributions $p_{\mathcal{S}}$ and $p_{\mathcal{T}}$. 
This closed-form solution has motivated the definition of SWD 
using the slice sampling technique~\cite{neal2003slice}. The idea is  to select a $1$-dimensional subspace and  project two $d$-dimensional   distributions on the subspace to generate their marginal $1-$dimensional distributions and define SWD by integrating the $1-$dimensional WD  between the marginal distributions the  over all possible $1-$dimensional  subspaces. 
A one-dimensional slice for a distribution $p_\mathcal{S}$  is defined as:
\begin{equation}
\mathcal{R}p_\mathcal{S}(t;\bm{\gamma})=\int_{\mathcal{S}^{d-1}} p_\mathcal{S}(\bm{x})\bm{\delta}(t-\langle\bm{\gamma}, \bm{x}\rangle)d\bm{x},
\label{eq:radon}
\end{equation}
where $\bm{\delta}(\cdot)$ denotes the Kronecker delta function,  $\langle \cdot ,\cdot\rangle$ denotes the   inner dot product, $\mathbb{S}^{d-1}$ is the $d$-dimensional unit sphere, and $\bm{\gamma}$ is the one-dimensional projection direction.  The SWD then is defined as the following integral:
\begin{eqnarray}
SW(p_\mathcal{S},p_\mathcal{T})=   \int_{\mathbb{S}^{d-1}} W(\mathcal{R} p_\mathcal{S}(\cdot;\gamma),\mathcal{R} p_\mathcal{T}(\cdot;\gamma))d\gamma.
\label{eq:radonSWDdistance}
\end{eqnarray}
 
The main advantage of using the SWD for UDS is that   unlike WD,the integrand of \eqref{eq:radonSWDdistance} has a closed form solutio  for a known $\gamma$.
To computer the integral in Eq.~\eqref{eq:radonSWDdistance}, we can  use a Monte Carlo style integration. First, we sample the projection subspace $\bm{\gamma}$ from a uniform distribution, defined over the unit sphere,  compute the $1-$dimensional WD, and then approximate  Eq.~\eqref{eq:radonSWDdistance}  by computing the arithmetic mean over a sufficient number of random projection:
\begin{equation}
\hat{D}(p_\mathcal{S},p_\mathcal{T})\approx \frac{1}{L}\sum_{l=1}^L \sum_{i=1}^M| \langle\gamma_l, \phi(\bm{x}_{s_l[i]}^\mathcal{S}\rangle)- \langle\gamma_l, \phi(\bm{x}_{t_l[i]}^\mathcal{T})\rangle|^2
\label{eq:SWDempirical}
\end{equation}
where $\gamma_l\in\mathbb{S}^{f-1}$ is a uniformly drawn random sample from the unit $f$-dimensional ball $\mathbb{S}^{d-1}$, and  $s_l[i]$ and $t_l[i]$ are the sorted indices of $\{\gamma_l\cdot\phi(\bm{x}_i)\}_{i=1}^M$ for source and target domains, respectively. Following ~\cite{gabourie2019learning}, we rely on Eq.~\eqref{eq:SWDempirical} to compute SWD empirically.   We use  SWD  in our work  because:
 i)  as discussed, is a suitable metric for solving optimization problems of deep learning, ii) it can be computed efficiently due to its closed form solution, and iii) its empirical version can be computed efficiently according to \eqref{eq:SWDempirical}. 
 Many UDA   methods are developed based on variations of ~\eqref{eq:smallmainPrMatch} using different probability metrics and additional regularization  to enforce a helpful property. In our work, we aim for improving upon the previous work that obtain a model by solving ~\eqref{eq:smallmainPrMatch}  through inducing larger margins between the learned class clusters in the embedding space to mitigate the effect of domain shift in the target domain. In Figure~\ref{figMUDA:1}, we have visualized conceptually the positive effect of large interclass  margins between   classes which makes feature representations more compact.

\section{Proposed Algorithm for Increased Interclass Margins}

 We rely on the internally learned distribution to increase interclass margins. Following the explained rationale above, this distribution is  a multi-modal distribution $p_J( \cdot)=\phi_{\bm{v}}(p_{\mathcal{S}}( \cdot))$  with $k$ modes in $\mathcal{Z}$ (see Figure~\ref{figMUDA:1}, left). Note that formation of a multimodal distribution is not guaranteed but if we model the embedding space using network responses just prior to the eventual softmax layer, it is a prerequisite for our model to learn the source domain.
  In Figure~\ref{figMUDA:1} (left-top), we can see the margins between classes corresponds to
the geometric distances between the boundaries of modes of the source-learned internal distribution $\phi_{\bm{v}}(p_{\mathcal{S}}( \cdot))$. As seen in Figure~\ref{figMUDA:1} (left-bottom), domain shift is a result of deviations of the internal distribution for the target domain  $\phi_{\bm{v}}(p_{\mathcal{T}}( \cdot))$ from the source-learned internal distribution $\phi_{\bm{v}}(p_{\mathcal{S}}( \cdot))$ which leads to more overlap between the class clusters in the target domain. In other words, the effect of domain shift in the input space translates into cross the boundary between some of the class clusters.
Our idea is to  develop a mechanism to repulse the target domain data samples from interclass margins towards the class means, as visualized in Figure~\ref{figMUDA:1} (right-top). As a result, we intuitively  expect more model robustness with respect to domain shift in the input space, as visualized in Figure~\ref{figMUDA:1} (right-bottom). In other words, our goal is to make the class clusters more compact in the embedding space for the source domain to allow more variability in the target domain. To implement this idea, we  first estimate the internally learned distribution in $\mathcal{Z}$ as a parametric GMM distribution as follows:
\begin{equation}
{
p_J(\bm{z})=\sum_{j=1}^k \alpha_j
\mathcal{N}(\bm{z}|\bm{\mu}_j,\bm{\Sigma}_j),
}
\end{equation}  
where  $\bm{\mu}_j$ and $\bm{\Sigma}_j$ denote the mean   and co-variance matrices for each mode and $\alpha_j$ denotes the mixture weights. This probability distribution is a suitable model for the internal probability distribution because we know $k$.
Additionally, as opposed to the general GMM estimation problem in which we need to rely on iterative and time-consuming procedures, e.g., expectation maximization (EM)~\cite{moon1996expectation}, estimating the GMM parameters is simple our UDA setting.  Because the source data points are labeled and also $k$ is known. Hence,  we can estimate $\bm{\mu}_j$ and $\bm{\Sigma}_j$  specifically as parameters of $k$ independent Gaussian distributions. The mixture  weights  $\alpha_j$ can be computed easily by a Maximum a Posteriori (MAP) estimate. Let $\bm{S}_j$ denotes the training data points that belong to the class $j$, i.e., $\bm{S}_j=\{(\bm{x}_i^s,\bm{y}_i^s)\in \mathcal{D}_{\mathcal{S}}|\arg\max\bm{y}_i^s=j \}$.   Then,  the  MAP estimation for the GMM parameters can be computed as:
\begin{equation}
{
\begin{split}
&\hat{\alpha}_j = \frac{|\bm{S}_j|}{N},\hspace{2mm}\hat{\bm{\mu}}_j = \sum_{(\bm{x}_i^s,\bm{y}_i^s)\in \bm{S}_j}\frac{1}{|\bm{S}_j|}\phi_v(\bm{x}_i^s),\hspace{2mm}  \hat{\bm{\Sigma}}_j =\sum_{(\bm{x}_i^s,\bm{y}_i^s)\in \bm{S}_j}\frac{1}{|\bm{S}_j|}\big(\phi_v(\bm{x}_i^s)-\hat{\bm{\mu}}_j\big)^\top\big(\phi_v(\bm{x}_i^s)-\hat{\bm{\mu}}_j\big).
\end{split}
}
\label{eq:MAPestICLR}
\end{equation}    
As opposed to the high computational complexity of EM~\cite{roweis1998algorithms}, ~\eqref{eq:MAPestICLR} does not add a significant computational overload to perform UDA. If the source domain dataset is balanced, we only need to check whether data points $\bm{x}_i^s$ belong to  $\bm{S}_j$ to compute $\alpha_j$ which has the computational complexity of $O(N)$. If we denote the dimension of the embedding space with $F$, the computational  Complexity of computing $\bm{\mu}_j$ would be $O(NF/k)$. The computational complexity for computing the co-variance matrices $\bm{\Sigma}_j$ would $O(F(\frac{N}{k})^2)$. Finally, since there are $k$ classes, the total computational complexity for estimating the GMM distribution would be $O(\frac{FN^2}{k})$. If $O(F)\approx O(k)$, i.e., a reasonable assumption when modeling the embedding space as network responses in the final layer, then the  total computational complexity would be $O(N^2)$. This is a small computational overload compared to computational complexity of one epoch of  back-propagation which additionally  needs to be performed several times. Because  deep neural networks generally have significantly larger number of weights than the number of data points $N$.

 \begin{algorithm}[t]
\caption{IMUDA\label{IJCAI2021Alg}} 
 { 
\begin{algorithmic}[1]
 
\STATE \textbf{Input:}   The datasets $\mathcal{D}_{\mathcal{S}}=(\bm{X}_{\mathcal{S}},  \bm{Y}_{\mathcal{T}})$, $\mathcal{D}_{\mathcal{T}}=(\bm{X}_{\mathcal{S}} )$
\STATE \hspace{4mm}\textbf{Pretraining on the Source Domain:}
\STATE \hspace{4mm} $\hat{ \theta}_0=(\hat{\bm{w}}_0,\hat{\bm{v}}_0)  =\arg\min_{\theta}\sum_i \mathcal{L}(f_{\theta}(\bm{x}_i^s),\bm{y}_i^s)$
\STATE \hspace{2mm}  \textbf{GMM Estimation:}
\STATE \hspace{4mm} Use \eqref{eq:MAPestICLR} and estimate GMM paramters $\alpha_j, \bm{\mu}_j,$   $\Sigma_j$
\STATE \textbf{Domain Adaptation}: 
\STATE \hspace{2mm} \textbf{Pseudo-Dataset Generation:} 
\STATE \hspace{4mm} Generate $\mathcal{\hat{D}}_{\mathcal{P}}$ based on ~\eqref{eq:pesudosamples}
\FOR{$itr = 1,\ldots, ITR$ }
\STATE draw random data batches from $\mathcal{ D}_{\mathcal{S}}$, $\mathcal{ D}_{\mathcal{T}}$, and $\mathcal{ D}_{\mathcal{P}}$ 
  and update the model based on   ~\eqref{eq:mainPrMatchICLR}
\ENDFOR
\end{algorithmic}}
\label{algSAforUDA}
\end{algorithm} 

We use the estimated GMM distribution induce larger interclass margins to improve the solution by ~\eqref{eq:smallmainPrMatch}. 
To induce repulsive biases from the class margins (see Figure~\ref{figMUDA:1}, top-right), we first generate a pseudo-dataset in the embedding space with confident labels $\mathcal{D}_{\mathcal{P}}=(\textbf{Z}_{\mathcal{P}},\textbf{Y}_{\mathcal{P}})$  using randomly drawn samples from  the estimated GMM, where $\bm{Z}_{\mathcal{P}}=[\bm{z}_1^p,\ldots,\bm{z}_{N_p}^p]\in\mathbb{R}^{p\times N_p}, \bm{Y}_{\mathcal{P}}=[\bm{y}^p_1,...,\bm{y}^p_{N_p}]\in \mathbb{R}^{k\times N_p}$, and $\bm{z}_i$ is drawn randomly $\bm{z}_i^p\sim \hat{p}_J(\bm{z})$. To ensure that these samples lie away from the margins and within  the proximity of the corresponding class means in the embedding space, we feed all the initially drawn samples into the classifier subnetwork and include only the samples for which the classifier confidence about the predicted label is more than  a predetermined confidence threshold $0<\tau<1$. This procedure helps us to avoid including samples close to the margins. More specifically, we follow:
\begin{equation}
{
\begin{split}
\mathcal{D}_{\mathcal{P}}=\bigg\{&(\bm{z}_i^p,\bm{y}^p_i)|   \bm{z}_i^p\sim \hat{p}_J(\bm{z}),  \max\{ h(\bm{z}_i^p)\}>\tau, \bm{y}^p_i=\arg\max_i\{ h(\bm{z}_i^p)\}\bigg\}.
\end{split}
}
\label{eq:pesudosamples}
\end{equation}    
Sample selection based on the threshold values close to one, $\tau\approx 1$,   implies that the pseudo-dataset samples are in the proximity of the class means (see Figure~\ref{figMUDA:1}  (right-top)). This means that the margins between the empirical data clusters are larger in the generated pseudo-dataset compared to the source domain data empirical clusters in the embedding space. We benefit from this property of the pseudo-dataset and update ~\eqref{eq:smallmainPrMatch} to induce larger margins:
\begin{equation}
{
\begin{split}
\hat{\bm{v}},\hat{\bm{w}}=&\arg\min_{\bm{v},\bm{w}}\Big\{ \frac{\lambda}{N}\sum_{i=1}^N \mathcal{L}\big(h_{\bm{w}}(\phi_{\bm{v}}(\bm{x}_i^s)),\bm{y}_i^s\big) +\frac{\lambda}{N_p}\sum_{i=1}^{N_p} \mathcal{L}\big(h_{\bm{w}}(\bm{z}_i^p),\bm{y}_i^p\big) +  \hat{D}\big(\phi_{\bm{v}}(\bm{X}_{\mathcal{T}}),\bm{X}_{\mathcal{P}} \big) +  \hat{D}\big(\phi_{\bm{v}}(\bm{X}_{\mathcal{S}}), \bm{X}_{\mathcal{P}} \big)\Big\} ,
\end{split}
}
\label{eq:mainPrMatchICLR}
\end{equation}  
where $\hat{D}(\cdot,\cdot)$ denotes the empirical SWD probability distribution metric.  The first and the second terms in \eqref{eq:mainPrMatchICLR} are ERM terms for the source dataset and the pseudo-dataset to keep the embedding space discriminative. The third term is an alignment term that matches the source domain  distribution  with   the pseudo-dataset empirical distribution. The fourth term is an  alignment term that matches the target domain  distribution  with   the pseudo-dataset empirical distribution which as we describe possesses larger margins. These terms help to increase the interclass margins which as we described would increase the model generalizability.  Our proposed algorithm,  called  Increased Margins for Unsupervised Domain Adaptation (IMUDA), is presented  and visualized in Algorithm~\ref{algSAforUDA} and Figure~\ref{figMUDA:1}, respectively.

 \section{Empirical Validation}

 \subsection{Datasets and Tasks}
 We validate our method on four standard UDA benchmarks.

 \textbf{Digit recognition   tasks:} the three  MNIST ($\mathcal{M}$),
USPS ($\mathcal{U}$), and
SVHN ($\mathcal{S}$) are used as domains. Following   the   literature,   the UDA tasks include three digit recognition    tasks: $\mathcal{M}\rightarrow \mathcal{U}$, $\mathcal{U}\rightarrow \mathcal{M}$, and $\mathcal{S}\rightarrow \mathcal{M}$. we resized the images of SVHN dataset to $28\times 28$ images to have the same size of the MNIST and the USPS datasets.

\textbf{Office-31 Detest:} this dataset is probably the most common UDA dataset that consists of 31 visual classes with a total of  $4,652$ images. There are three domains: Amazon ($\mathcal{A}$), Webcam ($\mathcal{W}$) and DSLR ($\mathcal{D}$) with six  definable UDA tasks, defined in a pair-wise manner among the three domains.

\textbf{ImageCLEF-DA Dataset:} it consists of the 12 shared   classes between the Caltech-256 ($\mathcal{C}$), the   ILSVRC 2012 ($\mathcal{I}$), and the Pascal VOC 2012 ($\mathcal{P}$) visual recognition datasets as domains. This dataset is completely balanced as each class has 50 images or 600 images per domain.  
Since the domains and the classes have the same number of images, this dataset is a complements  the Office-31 dataset which has varying domain and class sizes. Similarly, there   six possible UDA tasks can be defined.


\textbf{VisDA-2017:}  the goal  is to train a model for  natural images by pretraining the model on samples of a synthetic domain and then adapt it to generalize   on the real image domain.  The synthetic images are generated using 3D models of objects  based on applying different lightning conditions across 12 classes. The dataset is larger with 280K images.

\subsection{Backbone Structure and Evaluation Protocol:}

We follow the   literature for the network structures for each dataset to make fair evaluation of our work against existing works possible.   The VGG16 network  is used as the backbone model for the digit recognition tasks.  For the Office-31, the ImageCLEF-DA  datasets, we use  the  50 network as the backbone. For the VisDa2017 dataset, we use the ReNet-101 network as the backbone. The backbone models are pretrained on the ImageNet dataset and their output is fed into a hidden layer with size 128, followed by the last year according to the number of classes in each dataset. The last layer is set to be a softmax layer and the embedding space $\mathcal{Z}$ is set to be the features before softmax. 
 .
We use the cross-entropy loss as the discrimination loss.  At each   training epoch, we computed the combined loss function on the training split of datasets of both domains and stopped training when the training loss function became stable.  We used Keras for implementation and   ADAM optimizer to solve the UDA optimization problems. We have used 100 projection to compute the SWD loss. We tune the learning rate for each dataset such that the training loss function reduces smoothly.  We have run our code on a cluster node equipped with 4 Nvidia Tesla P100-SXM2 GPU's.   We used the  classification rate on  the  testing set to measure performance of the algorithms.   We performed 5 training trials   and reported the average performance and the standard deviation on the testing sets for these trials. 
Our code is available as a supplement.

  For each task, we report the source-trained model performance (Source Only) on the target domain  as a baseline. Performance improvements over this baseline serves as a simple ablative study and demonstrates the positive effect of model adaptation.  We then adapt the model using  IMUDA algorithm and report the performance on the target domain. In our results, we report the average classification rate  and the standard deviation on the target domain,  computed over ten randomly initialized runs  for all datasets, except VisDA2017 for which we have reported the best result.
   Following our theorem, we set $\tau= 0.95$.  We also set   $\lambda=10^{-2}$. The selection process for these  values will be explored further.


There are many existing UDA methods in the literature which makes extensive comparison challenging. For our purpose, we need to select a subset of the these works. We included both pioneer and recent works to be representative of the recent progress in the field and also to present the improvements made over the pioneer works. We selected the methods that reported results on the majority of the benchmarks we used.  These  methods include those   based on adversarial learning: GtA~\cite{sankaranarayanan2018generate}, DANN~\cite{ganin2016domain}, SymNets~\cite{zhang2019domain} ADDA~\cite{tzeng2017adversarial}, MADA~\cite{pei2018multi},   CDAN~\cite{long2018conditional},  DMRL~\cite{wu2020dual},   DWL~\cite{xiao2021dynamic},   HCL~\cite{huang2021model}, and CGDM~\cite{du2021cross} and   the methods which are based on direct distribution matching:  DAN~\cite{long2015learning}, DRCN~\cite{ghifary2016deep}, RevGrad~\cite{ganin2014unsupervised},
 JAN~\cite{long2017deep},   JDDA~\cite{chen2019joint},   CADA-P~\cite{kurmi2019attending},   ETD~\cite{li2020enhanced},   MetaAlign~\cite{wei2021metaalign}, and FixBi~\cite{na2021fixbi}. For each benchmark dataset that we report our results, we included results of the above works if the original paper has reported performance on that dataset. In our Tables, bold font denotes the best performance among all methods. We report the pre-adaptation baseline performance  in the first row of each table, followed by UDA methods based on adversarial learning, then followed by UDA methods based on direct matching. We report our result in the last rows.

\subsection{Results}

We  have reported performances on the three digit recognition tasks  in Table~\ref{table:tabDA1}. We observe that IMUDA  is quite competitive in these tasks. We also note that ETD, JDDA, and ETD, DWL which use secondary alignment mechanisms perform competitively.  This observation suggests that using secondary alignment mechanisms is essential to improve current UDA methods given the performance levels of existing UDA methods.

We have provided the comparison result for the UDA tasks of  the Office-31 dataset  in Table~\ref{table:tabDA2}. We observe that IMUDA   on average leads to  the best performance result. IMUDA also leads  to the best results on two of the tasks and is quite competitive on the remaining   tasks. This is likely because the ``source only performance'' for both of the $\mathcal{D}\rightarrow \mathcal{W}$  
and $\mathcal{W}\rightarrow \mathcal{D}$ tasks are quite high, almost 100\%. This means that the domain gap between these two tasks is relatively small to begin with, i.e., the two distributions are highly matched prior to model adaptation. As a result, this observation intuitively concludes that inducing larger margins is not going to be very helpful because the interclass margins are already relatively large in the target domain and are similar to the source domain. Hence, the margins are not violated prior to   adaptation and we cannot benefit much from our algorithm to increase the margins further.  

Results for   the ImageCLEF-DA dataset are presented in Table~\ref{table:tabDA3}. We see that  IMUDA   leads to a significant performance boost on this dataset, compared to prior work. We observe that we have about 7\%  boost on average  compared to the next best performance.   This performance is likely because the ImageCLEF-DA dataset is fully balanced in terms of   the number of data points across both the  domains and the classes. As a result, matching the   internal distributions with a GMM distribution is more accurate.
Because we have used the empirical distributions   for domain alignment, a balanced source dataset makes the empirical source distribution more representative of the true source distribution in     ImageCLEF-DA.
The empirical interclass margin also has the same meaning across the domains because of data balance. We conclude that having a balanced training dataset in the source domain  can boost our  performance. One area for improving our algorithm is to replicate similar amount of improvement when the training dataset is imbalanced.

  \begin{table*}[t!]
 \setlength{\tabcolsep}{3pt}
 \centering 
{\small
\begin{tabular}{lc|ccc|c|ccc}   
\multicolumn{2}{c}{Method}    & $\mathcal{M}\rightarrow\mathcal{U}$ & $\mathcal{U}\rightarrow\mathcal{M}$ & $\mathcal{S}\rightarrow\mathcal{M}$ &Method & $\mathcal{M}\rightarrow\mathcal{U}$ & $\mathcal{U}\rightarrow\mathcal{M}$ & $\mathcal{S}\rightarrow\mathcal{M}$\\
\hline
\multicolumn{2}{c|}{GtA~\cite{sankaranarayanan2018generate}}& 92.8  $\pm$  0.9	&	90.8  $\pm$  1.3	&	92.4  $\pm$  0.9 &CDAN~\cite{long2018conditional} &93.9 &96.9& 88.5 \\
\multicolumn{2}{c|}{ADDA~\cite{tzeng2017adversarial}}& 89.4  $\pm$  0.2&90.1  $\pm$  0.8&76.0  $\pm$  1.8& ETD~\cite{li2020enhanced} & 96.4  $\pm$  0.3&96.3  $\pm$  0.1&97.9  $\pm$  0.4\\  
\multicolumn{2}{c|}{DWL~\cite{xiao2021dynamic}}& \textbf{97.3} &97.4& \textbf{98.1}&   &  & & \\  
\hline
\multicolumn{2}{c|}{RevGrad~\cite{ganin2014unsupervised}}&	 77.1  $\pm$  1.8 	&	73.0  $\pm$  2.0 	&	73.9  &JDDA~\cite{chen2019joint} & -& $97.0$ $\pm$0.2 & 93.1$\pm$0.2   \\ 
\multicolumn{2}{c|}{DRCN~\cite{ghifary2016deep}}&	 91.8  $\pm$  0.1 	&	73.7  $\pm$  0.4 	&	82.0  $\pm$  0.2 &  DMRL~\cite{wu2020dual}& 96.1 & \textbf{99.0} &96.2   \\
\hline
\multicolumn{2}{c|}{Source Only}&	 90.1$\pm$2.6	&	80.2$\pm$5.7	&	67.3$\pm$2.6   & Ours &    96.6  $\pm$  0.4	&	 98.3  $\pm$  0.3	& 96.6   $\pm$  0.9 \\
\end{tabular}}
\caption{ Performance comparison for UDA tasks between MINIST, USPS, and SVHN    datasets.    }
\label{table:tabDA1}
 \end{table*}

  \begin{table*}[t!]
 \centering 
{\small
\begin{tabular}{lc|cccccc|c}   
\multicolumn{2}{c}{Method}    & $\mathcal{A}\rightarrow\mathcal{W}$ & $\mathcal{D}\rightarrow\mathcal{W}$ & $\mathcal{W}\rightarrow\mathcal{D}$ &$\mathcal{A}\rightarrow\mathcal{D}$ &$\mathcal{D}\rightarrow\mathcal{A}$ &$\mathcal{W}\rightarrow\mathcal{A}$ & Average\\
\hline
\multicolumn{2}{c|}{Source Only~\cite{he2016deep}}&68.4  $\pm$  0.2 & 96.7  $\pm$  0.1&  99.3  $\pm$  0.1&  68.9  $\pm$  0.2 & 62.5  $\pm$  0.3&  60.7  $\pm$  0.3& 76.1  \\
\hline
\multicolumn{2}{c|}{GtA~\cite{sankaranarayanan2018generate}}&89.5  $\pm$  0.5& 97.9  $\pm$  0.3& 99.8  $\pm$  0.4& 87.7  $\pm$  0.5& 72.8  $\pm$  0.3& 71.4  $\pm$  0.4& 86.5 \\
\multicolumn{2}{c|}{DANN~\cite{ganin2016domain}}&   82.0  $\pm$  0.4& 96.9  $\pm$  0.2& 99.1  $\pm$  0.1& 79.7  $\pm$  0.4& 68.2  $\pm$  0.4 &67.4  $\pm$  0.5 &82.2 \\ 
\multicolumn{2}{c|}{ADDA~\cite{tzeng2017adversarial}}& 86.2  $\pm$  0.5 & 96.2  $\pm$  0.3&  98.4  $\pm$  0.3&  77.8  $\pm$  0.3 & 69.5  $\pm$  0.4&  68.9  $\pm$  0.5&82.8 \\
\multicolumn{2}{c|}{SymNets~\cite{zhang2019domain}}& 90.8  $\pm$  0.1& 98.8  $\pm$  0.3& \textbf{\textbf{100}.0}  $\pm$  .0& 93.9  $\pm$  0.5& 74.6  $\pm$  0.6& 72.5  $\pm$  0.5 & 88.4\\  
\multicolumn{2}{c|}{MADA~\cite{pei2018multi}}& 82.0  $\pm$  0.4&  96.9  $\pm$  0.2 & 99.1  $\pm$  0.1&  79.7  $\pm$  0.4 & 68.2  $\pm$  0.4&  67.4  $\pm$  0.5  &82.2 \\
\multicolumn{2}{c|}{CDAN~\cite{long2018conditional} }&93.1  $\pm$  0.2 &98.2  $\pm$  0.2& \textbf{\textbf{100}.0}  $\pm$  0.0& 89.8  $\pm$  0.3& 70.1 $\pm$  0.4& 68.0  $\pm$  0.4 &86.6\\  
\multicolumn{2}{c|}{DMRL~\cite{wu2020dual} }& 90.8$\pm$0.3&99.0$\pm$0.2&\textbf{100}.0$\pm$0.0&93.4$\pm$0.5&73.0$\pm$0.3 &   71.2$\pm$0.3&87.9 \\ 
\multicolumn{2}{c|}{DWL~\cite{xiao2021dynamic} }& 89.2& 99.2 & \textbf{100.0}& 91.2& 73.1& 69.8& 87.1 \\ 
\hline
\multicolumn{2}{c|}{DAN~\cite{long2015learning}}&  80.5 $\pm$ 0.4 &97.1 $\pm$ 0.2& 99.6 $\pm$ 0.1& 78.6 $\pm$ 0.2& 63.6 $\pm$ 0.3& 62.8 $\pm$ 0.2&80.4\\ %
\multicolumn{2}{c|}{DRCN~\cite{ghifary2016deep}}&	 72.6  $\pm$  0.3 	&	96.4  $\pm$  0.1	& 99.2  $\pm$  0.3 	& 67.1  $\pm$  0.3 	& 56.0  $\pm$  0.5 	&$72.6$ $\pm$  0.3 & 77.7	  \\ 
\multicolumn{2}{c|}{RevGrad~\cite{ganin2014unsupervised}} &82.0  $\pm$  0.4&96.9  $\pm$  0.2& 99.1  $\pm$  0.1& 79.7  $\pm$  0.4& 68.2  $\pm$  0.4& 67.4  $\pm$  0.5 &82.2\\
\multicolumn{2}{c|}{CADA-P~\cite{kurmi2019attending}}&83.4$\pm$0.2 &99.8$\pm$0.1  & \textbf{100}.0$\pm$0   &80.1$\pm$0.1 &59.8$\pm$0.2 &59.5$\pm$0.3 &80.4\\ 
\multicolumn{2}{c|}{JAN~\cite{long2017deep}}& 85.4  $\pm$  0.3& 97.4  $\pm$  0.2& 99.8  $\pm$  0.2& 84.7  $\pm$  0.3& 68.6  $\pm$  0.3 &70.0  $\pm$  0.4 &84.3 \\ 
\multicolumn{2}{c|}{JDDA~\cite{chen2019joint}}&82.6  $\pm$  0.4& 95.2  $\pm$  0.2& 99.7  $\pm$  0.0 &79.8  $\pm$  0.1 &57.4  $\pm$  0.0& 66.7  $\pm$  0.2 &80.2\\ 
\multicolumn{2}{c|}{ETD~\cite{li2020enhanced}}&92.1&\textbf{100}.0 & \textbf{\textbf{100}.0}&88.0&71.0&67.8  86.2&   86.5\\ 
\multicolumn{2}{c|}{MetaAlign~\cite{wei2021metaalign}}& 93.0 $\pm$ 0.5& 98.6 $\pm$ 0.0& \textbf{100} $\pm$ 0.0& 94.5 $\pm$ 0.3& 75.0 $\pm$ 0.3& 73.6 $\pm$ 0.0& 89.2\\
\multicolumn{2}{c|}{HCL~\cite{huang2021model}}&92.5 &98.2& \textbf{100.0}& 94.7 & 75.9 &77.7& 89.8   \\
\multicolumn{2}{c|}{FixBi~\cite{na2021fixbi}}&96.1$\pm$0.2& 99.3$\pm$0.2& \textbf{100.0}$\pm$0.0 &95.0$\pm$0.4& \textbf{78.7}$\pm$0.5 &\textbf{79.4}$\pm$0.3& 91.4 \\
 \hline
\multicolumn{2}{c|}{IMUDA}&	 \textbf{99.6}  $\pm$  0.2	&	98.1  $\pm$  0.2	&	99.6  $\pm$  0.1 & \textbf{99.0}  $\pm$  0.4   & 74.9  $\pm$  0.4   & 78.7  $\pm$  1.1 & \textbf{91.7}\\   
\end{tabular}}
\caption{ Performance comparison  for UDA tasks for  Office-31 dataset. }
\label{table:tabDA2}
 \end{table*}

  \begin{table*}[t!]
 \centering 
{\small
\begin{tabular}{lc|ccccccc}   
\multicolumn{2}{c}{Method}    & $\mathcal{I}\rightarrow\mathcal{P}$ & $\mathcal{P}\rightarrow\mathcal{I}$ & $\mathcal{I}\rightarrow\mathcal{C}$ &$\mathcal{C}\rightarrow\mathcal{I}$ &$\mathcal{C}\rightarrow\mathcal{P}$ &$\mathcal{P}\rightarrow\mathcal{C}$& Average \\
\hline
\multicolumn{2}{c|}{Source Only~\cite{he2016deep}}& 74.8  $\pm$  0.3& 83.9  $\pm$  0.1& 91.5  $\pm$  0.3 &78.0  $\pm$  0.2 &65.5  $\pm$  0.3& 91.2  $\pm$  0.3 & 80.8 \\
\hline
\multicolumn{2}{c|}{DANN~\cite{ganin2016domain}}&   82.0  $\pm$  0.4& 96.9  $\pm$  0.2& 99.1 $\pm$  0.1& 79.7  $\pm$  0.4& 68.2  $\pm$  0.4 &67.4  $\pm$  0.5 &82.2 \\ 
\multicolumn{2}{c|}{SymNets~\cite{zhang2019domain}}& 80.2 $\pm$ 0.3 &93.6 $\pm$ 0.2& 97.0 $\pm$ 0.3 &93.4 $\pm$ 0.3 &78.7 $\pm$ 0.3 &96.4 $\pm$ 0.1&89.9 \\  
\multicolumn{2}{c|}{MADA~\cite{pei2018multi}}& 75.0 $\pm$ 0.3& 87.9 $\pm$ 0.2& 96.0 $\pm$ 0.3& 88.8 $\pm$ 0.3& 75.2 $\pm$ 0.2& 92.2 $\pm$ 0.3 &85.9   \\
\multicolumn{2}{c|}{CDAN~\cite{long2018conditional} }&76.7 $\pm$ 0.3& 90.6 $\pm$ 0.3& 97.0 $\pm$ 0.4 &90.5 $\pm$ 0.4& 74.5 $\pm$ 0.3 &93.5 $\pm$ 0.4 & 87.1\\ 
\multicolumn{2}{c|}{DMRL~\cite{wu2020dual} }&77.3$\pm$0.4&90.7$\pm$0.3&97.4$\pm$0.3&91.8$\pm$0.3&76.0$\pm$0.5&94.8$\pm$0.3& 88.0 \\ 
\multicolumn{2}{c|}{DWL~\cite{xiao2021dynamic} }&
82.3 &94.8& 98.1& 92.8& 77.9& 97.2 &90.5\\
\hline
\multicolumn{2}{c|}{DAN~\cite{long2015learning}}& 74.5 $\pm$ 0.4 &82.2 $\pm$ 0.2 &92.8 $\pm$ 0.2& 86.3 $\pm$ 0.4& 69.2 $\pm$ 0.4& 89.8 $\pm$ 0.4&82.4\\ 
\multicolumn{2}{c|}{RevGrad~\cite{ganin2014unsupervised}} &75.0 $\pm$ 0.6 &86.0 $\pm$ 0.3& 96.2 $\pm$ 0.4 &87.0 $\pm$ 0.5& 74.3 $\pm$ 0.5& 91.5 $\pm$ 0.6&85.0 \\
\multicolumn{2}{c|}{JAN~\cite{long2017deep}}&  76.8 $\pm$ 0.4&  88.0 $\pm$ 0.2&  94.7 $\pm$ 0.2 & 89.5 $\pm$ 0.3&  74.2 $\pm$ 0.3 & 91.7 $\pm$ 0.3&85.8\\
\multicolumn{2}{c|}{CADA-P~\cite{kurmi2019attending}} &78.0&90.5 & 96.7&92.0 & 77.2&  95.5 & 88.3\\\multicolumn{2}{c|}{ETD~\cite{li2020enhanced}} &81.0 & 91.7 & 97.9  &93.3 & 79.5  &95.0&89.7\\
\multicolumn{2}{c|}{CGDM~\cite{du2021cross}}&  78.7 $\pm$ 0.2& 93.3 $\pm$ 0.1& 97.5 $\pm$ 0.3& 92.7 $\pm$ 0.2& 79.2 $\pm$ 0.1& 95.7 $\pm$ 0.2& 89.5\\
\hline
\multicolumn{2}{c|}{IMUDA}&	 		\textbf{89.5}  $\pm$  1.2	&	\textbf{99.8}  $\pm$  0.2 & \textbf{\textbf{100}}  $\pm$  0.0   & \textbf{99.9}  $\pm$  0.1 & \textbf{92.6}  $\pm$  0.9 & \textbf{99.8}  $\pm$  0.2  &\textbf{96.9}    \\
\end{tabular}}
\caption{ Performance comparison    for UDA tasks for  ImageCLEF-DA dataset. }
\label{table:tabDA3}
 \end{table*}

  \begin{table*}[t!]
 \centering 
{\footnotesize
\begin{tabular}{lc|cccccccccccc|c}   
\multicolumn{2}{c}{Method}    & Plane& Bike &Bus &Car& Horse &Knife& Motor& Person &Plant &Skateboard& Train& Truck & Average\\
\hline
\multicolumn{2}{c|}{Source Only}&70.6& 51.8& 55.8& 68.9 &77.9& 7.6 &93.3 &34.5& 81.1 &27.9 &88.6 &5.6 &55.3  \\
\hline
\multicolumn{2}{c|}{DANN}&81.9 &77.7 &82.8 &44.3 &81.2 &29.5 &65.1 &28.6& 51.9 &54.6 &82.8 &7.8& 57.4 \\
\multicolumn{2}{c|}{RevGrad}& 75.9 &70.5 &65.3& 17.3 &72.8 &38.6 &58.0 &77.2 &72.5 &40.4 &70.4 &44.7 &58.6 \\
\multicolumn{2}{c|}{JAN}& 92.1&  66.4&  81.4 & 39.6 & 72.5 & 70.5 & 81.5&  70.5&  79.7 & 44.6 & 74.2 & 24.6 & 66.5 \\
\multicolumn{2}{c|}{GtA}&  - &- &- &- &- &- &- &- &- &- &- &- & 77.1 \\
\multicolumn{2}{c|}{CDAN}& 85.2&  66.9&  83.0&  50.8&  84.2& 74.9&  88.1&  74.5&  83.4&  76.0&  81.9 & 38.0&  73.7 \\
\multicolumn{2}{c|}{MCD}& 87.0 &60.9& 83.7 &64.0& 88.9 &79.6& 84.7 &76.9& 88.6 &40.3 &83.0& 25.8 &71.9 \\
\multicolumn{2}{c|}{DMRL}& - &- &- &- &- &- &- &- &- &- &- &- & 75.5\\
\multicolumn{2}{c|}{DAN}& 68.1 &15.4& 76.5& 87 &71.1& 48.9& 82.3 &51.5& 88.7 &33.2& 88.9 &42.2 &61.1 \\
\multicolumn{2}{c|}{DWL}&90.7& 80.2& 86.1& 67.6 &92.4 &81.5 &86.8 &78.0 &90.6 &57.1 &85.6 &28.7 &77.1  \\
\multicolumn{2}{c|}{CGDM}& 93.4 &82.7 &73.2 &68.4& 92.9 &94.5& 88.7& 82.1 &93.4& 82.5& 86.8& 49.2& 82.3 \\
\multicolumn{2}{c|}{HCL}& 93.3 &85.4& 80.7& 68.5& 91.0& 88.1& 86.0 &78.6& 86.6 &88.8 &80.0 &74.7& 83.5 \\
\multicolumn{2}{c|}{BiFix}& 96.1 &\textbf{87.8} &90.5 &\textbf{90.3}& \textbf{96.8} &95.3 &92.8& \textbf{88.7}& \textbf{97.2}& 94.2& 90.9& 25.7& \textbf{87.2}\\
 \hline
\multicolumn{2}{c|}{IMUDA}&	 \textbf{98.5}& 63.9& \textbf{92.8}& 74.9& 84.4& \textbf{98.8}& \textbf{93.9}& 86.1& 92.7& \textbf{95.5}& \textbf{94.2}&\textbf{45.3}& 85.1 \\   
\end{tabular}}
\caption{Performance    for the VisDA UDA task. }
\label{table:tabDA5}
 \end{table*}


 \begin{figure*}[t]
  \centering
    \begin{subfigure}[b]{0.24\textwidth}\includegraphics[width=\textwidth]{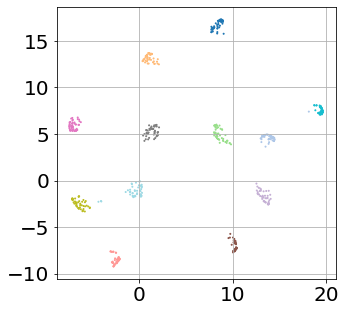}
        \caption{ }
        \label{fig11}
    \end{subfigure}
  \centering
    \begin{subfigure}[b]{0.24\textwidth}\includegraphics[width=\textwidth]{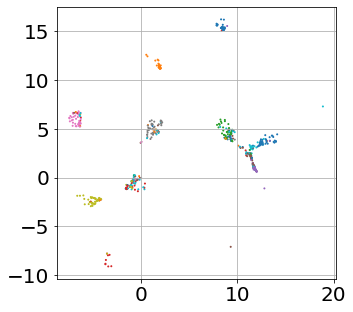}
        \caption{ }
        \label{fig12}
    \end{subfigure}
       \begin{subfigure}[b]{0.24\textwidth}\includegraphics[width=\textwidth]{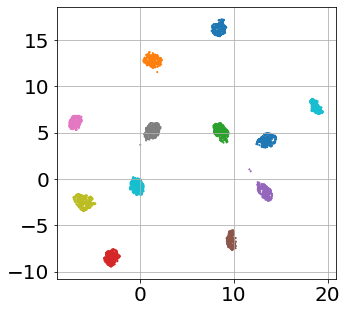}
           \centering
       \caption{ }
        \label{fig13}
    \end{subfigure}
      \centering
           \begin{subfigure}[b]{0.24\textwidth}\includegraphics[width=\textwidth]{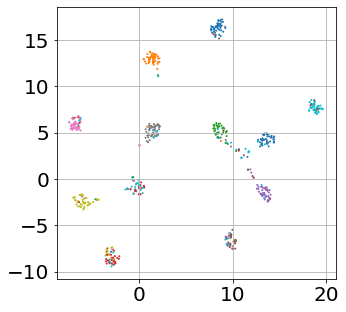}
           \centering
        \caption{ }
        \label{fig14}
    \end{subfigure}
     \caption{UMAP visualization for the representations of the dataset testing split for  the $\mathcal{C}\rightarrow \mathcal{P}$ task:  (a) the source domain (b) the target domain prior to adaptation, (c) samples drawn from the learned GMM, (d) the target domain after adaptation.  (Best viewed enlarged on screen and in color).  }\label{fig1T}
\end{figure*}

 \begin{figure*}[t]
  \centering
    \begin{subfigure}[b]{0.24\textwidth}\includegraphics[width=\textwidth]{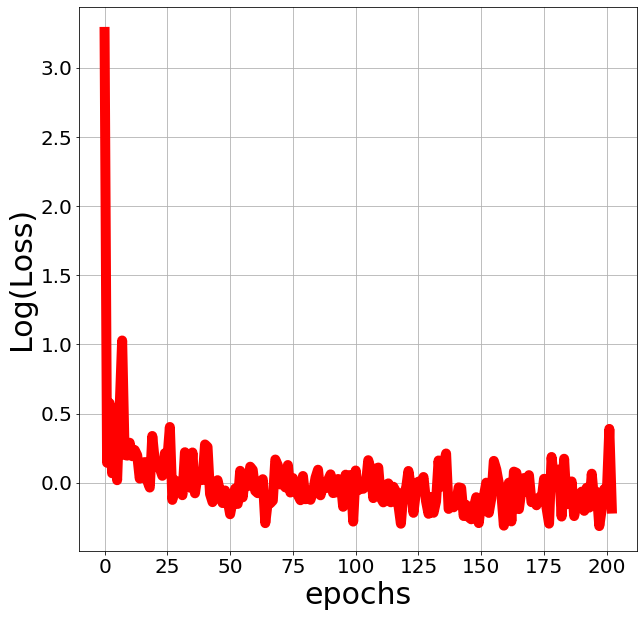}
        \caption{ }
        \label{fig21}
    \end{subfigure}
  \centering
    \begin{subfigure}[b]{0.22\textwidth}\includegraphics[width=\textwidth]{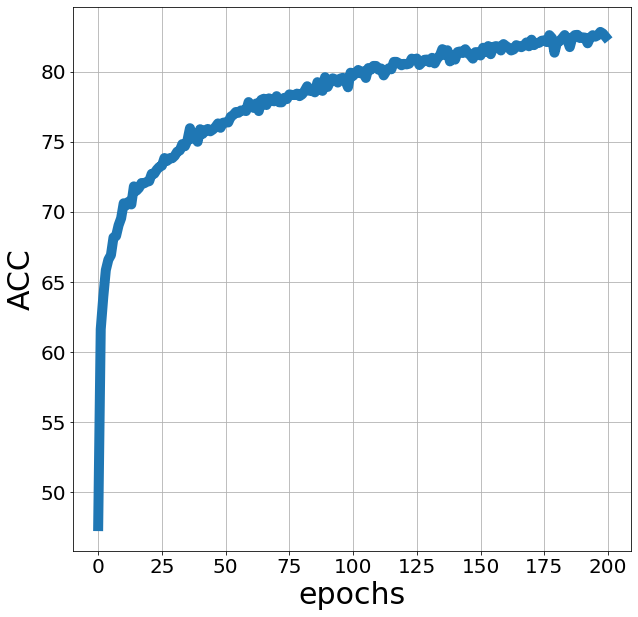}
        \caption{ }
        \label{fig22}
    \end{subfigure}
       \begin{subfigure}[b]{0.24\textwidth}\includegraphics[width=\textwidth]{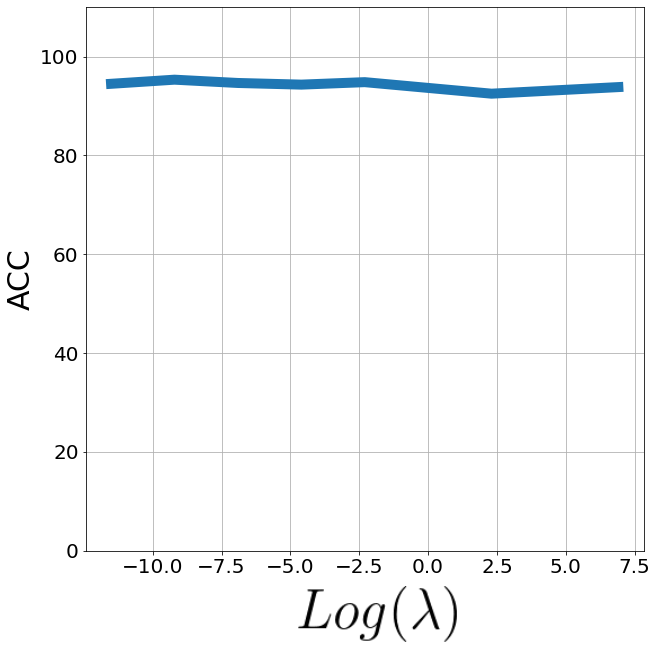}
           \centering
       \caption{ }
        \label{fig23}
    \end{subfigure}
      \centering
           \begin{subfigure}[b]{0.24\textwidth}\includegraphics[width=\textwidth]{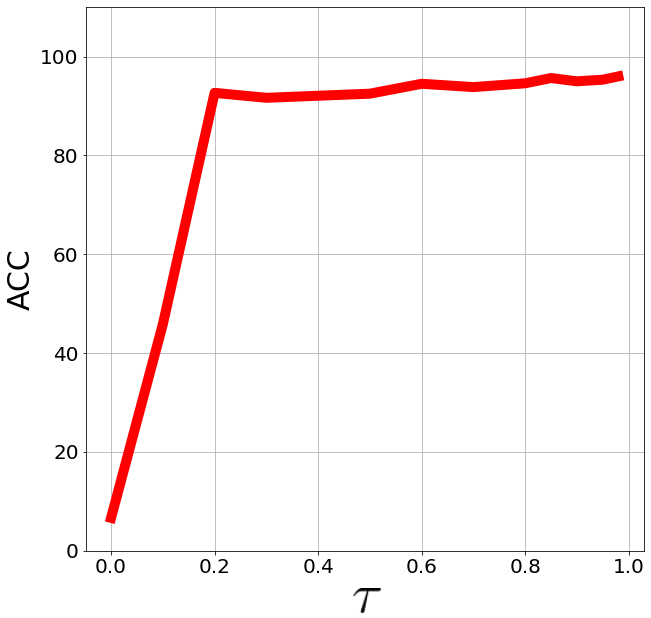}
           \centering
        \caption{ }
        \label{fig24}
    \end{subfigure}
     \caption{Empirical analysis based on  the VisDA task:  (a) loss function on the training split versus \#epochs and (b) learning curve for the testing split versus \#epochs; Effect of parameter values for the $\mathcal{C}\rightarrow \mathcal{P}$ task (c) performance versus the trade-off parameter $\lambda$ and (d) classification accuracy versus the confidence parameter $\tau$. (Best viewed enlarged on screen and in color).  }\label{fig2T}
\end{figure*}


We have included the performance results for the  single task of the VisDA2017 dataset in Table~\ref{table:tabDA5}.
We   observe a competitive performance on this dataset. The large size of this dataset, makes the empirical probability distribution a more accurate representation of the true distribution. For this reason, empirical SWD loss can enforce domain alignment  better. Quite intuitively, we conclude that having larger training datasets helps boosting our performance.

From Tables~1--4, we can conclude that despite simplicity, IMUDA  is   a competitive UDA method when compared against the existing works given that it either outperforms the existing methods or its performance is close to the best method.   Note that due to diversity of the benchmark UDA tasks, no single UDA method outperforms all the other existing UDA methods on all the major UDA tasks in the literature. Moreover, it is highly likely that performances for a given algorithm can change to some extend by finetuning the hyper-parameters/parameters and selecting different optimization techniques. Hence, those algorithms with close performances should be treated to have equally competitive performances.
We can see that each algorithms leads to the best performance only on a subset of the tasks.
 The diversity of experiments helped identifying circumstances under which, our algorithm likely would work better.

\subsection{Analytic and Ablative  Analysis}

We empirically analyzed our algorithm for further exploration and providing better understanding of its effect on data representation. We first checked the effect of the algorithm on the alignment of the distributions  in the embedding space. We used the $\mathcal{C}\rightarrow \mathcal{P}$ task of the ImageClef-DA dataset for this purpose. After passing the testing split of  the data for both domains into the embedding space, we used the UMAP~\cite{mcinnes2018umap} visualization tool to reduce  the dimension of the   data representations  to two for 2D visualization.
We have visualized the result in  Figure~\ref{fig1T}. We have visualized the    source domain testing split representations, samples that are drawn from  the estimated GMM distribution, and the target domains testing split representations, prior and after performing unsupervised domain adaptation.  In this figure, each point represents a single data point and each color represents one of the twelve classes. Comparing Figures~\ref{fig11}  and \ref{fig13}, we  observe that high-confidence GMM samples match the source domain distribution quite well for this task which suggests   GMM is a good parametric model for the source distribution.  Comparing Figure~\ref{fig12} with Figure~\ref{fig11}, we observe that domain shift has led to overlapping clusters in the target domain, similar to the qualitative visualization in Figure~\ref{figMUDA:1}. Finally,  Figure~\ref{fig14} demonstrate that the IMUDA algorithm successfully has aligned the distribution of the target domain with the distribution of the source domain,mitigating domain shift. This empirical observations supports the intuition behind our approach for developing the algorithm, visualized in Figure~\ref{figMUDA:1}.

For further investigation, we have plotted the training loss function value in ~\eqref{eq:smallmainPrMatch} as well as the classification performance on the testing split of the VisDA task versus  the number of optimization epochs in Figures~\ref{fig21} and ~\ref{fig22}, respectively. Comparing the two curves, we observe that as the   optimization objective loss function decreases, i.e., the distance between the two empirical distributions decreases, model generalization on the target domain increases. This observation  confirms that IMUDA algorithm implements the desired domain alignment effect for UDA and accords nicely with our reasoning that more domain alignment can lead to better performance on the target domain data.

We have studied the effect of the values of the primary hyper-parameters of the algorithm on UDA performance in Figure~\ref{fig23} and Figure~\ref{fig24}. An advantage of our algorithm over some of the existing UDA methods is that IMUDA only has two primary hyper-parameters: $\lambda$ and $\tau$. We have plotted the classification accuracy versus varying values of the hyper-parameter $\lambda$ in Figure~\ref{fig23}. We can see that the performance is relatively constant with respect to this parameter. This is expected because  the ERM terms in ~\eqref{eq:mainPrMatchICLR} are already small prior to the UDA optimization due to the pretraining step. Hence, when the UDA optimization is  performed, it mainly minimizes the alignment loss terms. Figure~\ref{fig24} presents the classification accuracy versus varying values of the confidence hyper-parameter $\tau$.  As predictable from prior discussion, we observe that larger values for $\tau$ improve the performance. Very Small values for $\tau$ can degrade the performance because the low-confidence GMM samples can potentially behave as outliers, making domain alignment even more challenging compared to simple UDA based on distribution matching. Quite importantly,
this experiments also serves as an ablation study to confirm that using high-confidence samples is indeed critical for our algorithm to improve model generalization. We can see as we use larger $\tau$, i.e., a larger interclass margin in the source domain, the UDA performance also increases. This observation conforms that our secondary mechanism to induce larger margin is indeed effective. We also observe that after $\tau\approx 0.2$, increasing $\tau$ is not as helpful. As we can see, this observation is likely because the performance is already high which means the margins are already high. Note, however, the performance still is increasing with a smaller slope for $\tau>0.2$.
 Finally, we have performed ablative experiments on the  optimization terms in ~\eqref{eq:smallmainPrMatch}. Note that our experiments on the parameter $\lambda$, already inlcuded the effect of dropping the $1^{st}$ and the $2^{nd}$ terms in ~\eqref{eq:smallmainPrMatch}. Hence,  we have studied effect of dropping the $3^{rd}$ and the $4^{th}$ terms    to study the effect of each term on performance using the Office-31 dataset. The results are presented in  Table~\ref{table:abe}. Comparing the performances with IMUDA, we observe that both terms contribute to the optimal performance, but the $3^{rd}$ term is more crucial. This observation is expected because the $3^{rd}$ term enforces alignment of the target domain with the pseudo-dataset directly. However, the $4^{th}$ term is helpful to induce larger margins. We conclude that the $4^{th}$ term helps to implement a complementary mechanism to improve the UDA  results. 
 
   \begin{table*}[t!]
 \centering 
{\small
\begin{tabular}{lc|cccccc|c}   
\multicolumn{2}{c}{Method}    & $\mathcal{A}\rightarrow\mathcal{W}$ & $\mathcal{D}\rightarrow\mathcal{W}$ & $\mathcal{W}\rightarrow\mathcal{D}$ &$\mathcal{A}\rightarrow\mathcal{D}$ &$\mathcal{D}\rightarrow\mathcal{A}$ &$\mathcal{W}\rightarrow\mathcal{A}$ & Average\\
\hline
\multicolumn{2}{c|}{$3^{rd}$ Term}&	  70.2 $\pm$ 0.9 & 91.3 $\pm$ 1.9 & 96.8 $\pm$    0.5&77.4 $\pm$ 0.4 & 58.1$\pm$0.3  & 57.7 $\pm$ 0.4 &  75.3\\
\multicolumn{2}{c|}{$4^{th}$ Term}&	  99.3 $\pm$ 0.3 & 95.1 $\pm$ 0.5 & 98.9 $\pm$    0.1& 99.0$\pm$ 0.1 &77.0 $\pm$0.5  & 77.8 $\pm$0.1  & 91.2 \\
\multicolumn{2}{c|}{IMUDA}&	 99.6  $\pm$  0.2	&	98.1  $\pm$  0.2	&	99.6  $\pm$  0.1 & 99.0  $\pm$  0.4   & 74.9  $\pm$  0.4   & 78.7  $\pm$  1.1 & 91.7\\   
\end{tabular}}
\caption{Ablative Experiments on Optimization Terms Using the  Office-31 dataset. }
\label{table:abe}
 \end{table*}

 \section{Conclusions  }
 In this paper, we developed a UDA algorithm by developing a mechanism that increases the interclass margins between the formed class clusters in an embedding space to reduce the effect of domain shift. The embedding space is  modeled by responses of a neural network in its final layer.  Increasing the interclass margins mitigates the effect of domain shift on the model generalization in the target domain. Our algorithm is based on learning the internal distribution of the source domain and use it to repulse the target domain data representations in the embedding space away from the class boundaries. We estimate this distribution as a parametric Gaussian mixture model (GMM).  We then draw random samples with confident labels from the GMM and use them to induce larger interclass margins.  Our  empirical results suggest that our approach is effective and compares favorably against existing methods. We also conclude that secondary mechanisms are necessary to improve domain alignment for  UDA given the current performance level of UDA algorithms.
 Future works includes performance improvement when the source dataset is imbalanced.

{

}

 \end{document}